# Sidewalk Hazard Detection Using Variational Autoencoder and One-Class SVM

Edgar Guzman
*Harvard School of Engineering and Applied Sciences*
Boston, United States
eguzman@g.harvard.edu

Robert D. Howe
*Harvard School of Engineering and Applied Sciences*
Boston, United States
howe@seas.harvard.edu

*Abstract*—The unpredictable nature of outdoor settings introduces numerous safety concerns, making hazard detection crucial for safe navigation. This paper introduces a novel system for sidewalk safety navigation utilizing a hybrid approach that combines a Variational Autoencoder (VAE) with a One-Class Support Vector Machine (OCSVM). The system is designed to detect anomalies on sidewalks that could potentially pose walking hazards. A dataset comprising over 15,000 training frames and 5,000 testing frames was collected using video recordings, capturing various sidewalk scenarios, including normal and hazardous conditions. During deployment, the VAE utilizes its reconstruction mechanism to detect anomalies within a frame. Poor reconstruction by the VAE implies the presence of an anomaly, after which the OCSVM is used to confirm whether the anomaly is hazardous or non-hazardous. The proposed VAE model demonstrated strong performance, with a high Area Under the Curve (AUC) of 0.94, effectively distinguishing anomalies that could be potential hazards. The OCSVM is employed to reduce the detection of false hazard anomalies, such as manhole or water valve covers. This approach achieves an accuracy of 91.4%, providing a highly reliable system for distinguishing between hazardous and non-hazardous scenarios. These results suggest that the proposed system offers a robust solution for hazard detection in uncertain environments.

## I. Introduction

Vision is an essential sense that helps perceive and interpret the environment, allowing navigation, decision-making, and interaction with the surroundings. As humans, we rely on vision to connect with our environment to perform fast and accurate tasks in complex settings. However, navigating through the world can be challenging for many, including the elderly, the visually impaired, and even robots. Thus, research in safety navigation has made great progress in the past decade, creating intuitive systems capable of generating trajectory plans and preventing collisions [1]–[3]. However, most of these systems have been implemented and evaluated for obstacle avoidance. For widespread adoption, a more comprehensive approach, where safety systems can detect hazards, is needed.

A promising approach to gathering information about the environment involves the use of inexpensive video cameras. These sensors are capable of detecting objects within each frame [4], potentially indicating the presence of a hazard. With the integration of artificial intelligence, detecting abnormalities has become significantly easier. Anomaly detection has been applied in surveillance systems to enhance the security of public facilities, transportation networks, and infrastructure [5]–[7]. Additionally, industrial applications are leveraging anomaly detection to identify unusual patterns or defects within their products [8].

In the context of autonomous safety navigation, commercial vehicles and robots have also incorporated the use of anomaly detection to identify irregular patterns within their trajectories [9]–[11]. While vision-based machine learning approaches have proven to enhance the detection of anomalies in many fields, their applications to safety navigation for the elderly or visually impaired have not advanced as rapidly.

Recent work for the visually impaired and assistive lowerlimb devices has integrated wearable RGB cameras to avoid hazardous situations, such as changes in terrain (i.e., staircases, ramps, curbs) [12]–[14]. Research also includes detecting hazards in traversal areas with a polarized RGB-D camera by identifying the polarized light reflected by puddles [15]. Additionally, some studies utilize an RGB-D depth module to identify potholes or other concave anomalies [16]. Another approach involves training a CNN to detect abnormalities within a walking trajectory [17]. However, these advancements are not without limitations: the first approach relies on adding polarized lenses to a camera that is only capable of detecting specularly reflective hazards. Similarly, the second study focuses on concave anomalies. The last approach uses a CNN that requires an extensive amount of data and variation of anomalies in the environment, which makes it impossible to capture all scenarios. Therefore, a robust system capable of detecting a wide range of anomalies without needing large datasets, sensor adjustments, or constraints to specific types of anomalies is needed.

In this paper, we present a hazard detection system designed for sidewalk navigation. Our approach utilizes a variational autoencoder fused with a one-class support vector machine (OCSVM) to process RGB data. The variational autoencoder attempts to regenerate an input image by compressing the input image into a lower-dimensional latent space and then reconstructs it. If thereconstruction probability error exceeds a predefined threshold, implying the presence of an anomaly, the latent vector generated by the variational autoencoder is passed to a OCSVM for a final evaluation to determine whether the anomaly is hazardous or not. The system can alert users when hazardous anomalies are detected in their walking trajectory by outlining the region where the abnormal object is located using a pixel-wise error function, thus warning of dangerous scenarios.

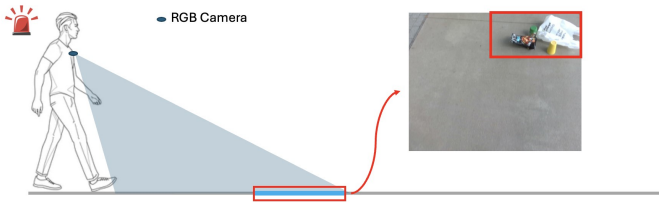

Fig. 1: Detection of hazard while walking.

In the next section, we describe the system's design and provide an overview. We then proceed to describe the methodology of our study, including the dataset utilized and the approaches employed across the autoencoder and the OCSVM. Subsequently, we present a section dedicated to the results, where we independently evaluate the effectiveness and accuracy of the autoencoder and OCSVM. The paper concludes with a discussion of the implications of our findings, addressing system-level considerations, and exploring potential directions for future research in this domain.

## II. SYSTEM DESIGN

### A. Overview

In the proposed system, the user wears a video camera as they navigate sidewalks, where RGB frames are passed through a Variational Autoencoder (VAE) that has been trained on sidewalks without hazards. The VAE attempts to encode and decode the input image. If the encodedecode error is low, the frames are considered non-hazardous. Otherwise, the VAE has identified an anomaly, which could potentially be hazardous. The VAE's latent vector is then passed to a OCSVM. The OCSVM uses a learned boundary between recognized and non-recognized anomalies to determine the presence of a hazard within the input frame. Recognized anomalies from the OCSVM are labeled as non-hazardous, while non-recognized anomalies from the OCSVM are labeled as hazardous. If a hazard is detected, the system creates a bounding box over the detected anomaly, alerting the user to the location of the hazard. Figs. 1 and 2 provide a high-level overview of the proposed systems flow as well as a visualization of the application.

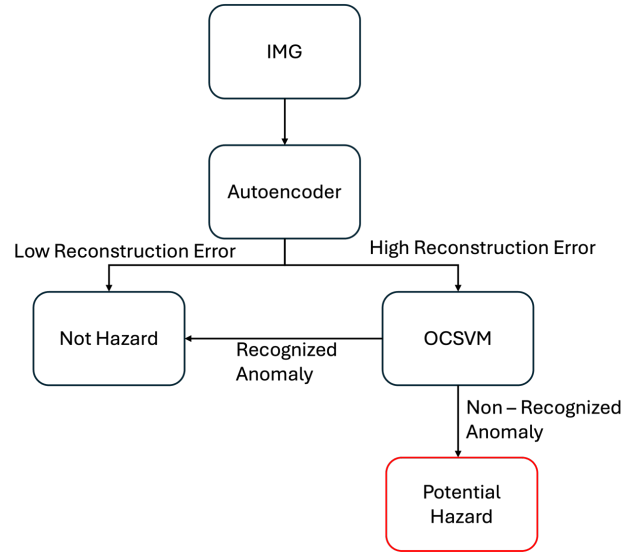

Fig. 2: Overview block diagram of the proposed approach.

### B. Data Collection

Normal sidewalk samples were collected from an RGB camera. The camera was placed at chest height, tilted approximately 60 degrees downward to view the sidewalk immediately in front of the user. Images of diverse locations featuring various types of sidewalks were captured, ensuring that the training data was generalized, allowing the system to function in different locations. Additionally, to enhance generalization, data augmentation was applied to adjust the brightness of the dataset, simulating conditions of both bright and cloudy days.

After normal data was collected, abnormal objects on various types of sidewalks were captured for testing purposes. Some of these anomalies were naturally occurring, like large cracks in the sidewalk, while others, such as litter, were artificially introduced to validate the concept.

The RGB camera used was an Intel RealSense D435i, providing an RGB field of view (FOV) of 87 x 58 with a 640 x 480 resolution, operating at 30 fps. The depth data from the camera was not used in this study.

### C. Variational Autoencoder

Given that we are using RGB frames as a method to detect the presence of anomalies, we employ a deep encoderdecoder architecture, commonly known as a Variational Autoencoder (VAE). VAEs are deep neural networks composed of an encoder, $F(x)$, where $x$ is the input. The encoder utilizes convolutional layers and ReLU activations (as shown in Fig. 4) to transform input images into a low-dimensional feature space, producing a mean vector $\mu$ and a standard deviation vector $\sigma$ as shown in Fig. 3, which define a Gaussian distribution $N(\mu,\sigma)$. A latent vector, $z$, is derived by randomly sampling $\epsilon$ from a standard normal Gaussian distribution ($\epsilon \sim N(0,1)$) and then mapping it back to $\mu$ and $\sigma$ such that $z = \mu + \sigma \times \epsilon$. This process, known as the

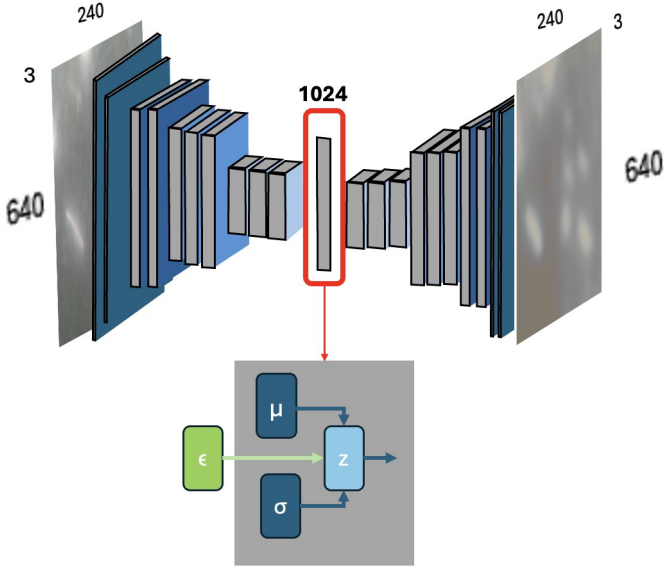

Fig. 3: Variational Autoencoder latent space reparameterization trick.

reparameterization trick, allows for stochastic sampling while keeping the model differentiable, enabling backpropagation during training. This method introduces stochasticity, which is advantageous compared to the deterministic output from traditional autoencoders because it allows the model to learn and generate a diverse set of latent representations. The decoder, $F'(z)$, also built with convolutional layers and ReLU activations (as shown in Fig. 4), then attempts to reconstruct the original image from the latent vector. For further details of VAE, see [18]

The VAE is trained to accurately capture the characteristics of the training data variability. During training, the VAE learns features from the input images and generates a latent vector specific to the training data. In this paper, the training data consists of variations of normal sidewalk images. As a result, during deployment, the VAE will struggle to produce an appropriate latent vector when presented with images containing anomalies, leading to a high reconstruction error.

```
----------------------------------------------------------------
        Layer (type)               Output Shape         Param #
================================================================
            Conv2d-1        [-1, 32, 160, 120]           2,432
              ReLU-2        [-1, 32, 160, 120]               0
            Conv2d-3         [-1, 64, 80, 60]           51,264
              ReLU-4         [-1, 64, 80, 60]                0
            Conv2d-5        [-1, 128, 40, 30]          204,928
              ReLU-6        [-1, 128, 40, 30]                0
            Conv2d-7        [-1, 256, 20, 15]          819,456
            Linear-8                [-1, 1024]       78,644,224
            Linear-9                [-1, 1024]       78,644,224
           Linear-10               [-1, 76800]       78,720,000
   ConvTranspose2d-11       [-1, 128, 40, 30]          524,416
             ReLU-12        [-1, 128, 40, 30]                0
   ConvTranspose2d-13        [-1, 64, 80, 60]          131,136
             ReLU-14         [-1, 64, 80, 60]                0
   ConvTranspose2d-15       [-1, 32, 160, 120]          32,800
             ReLU-16        [-1, 32, 160, 120]               0
   ConvTranspose2d-17         [-1, 3, 320, 240]          1,539
================================================================
Total params: 237,776,419
Trainable params: 237,776,419
Non-trainable params: 0
----------------------------------------------------------------
Input size (MB): 0.88
Forward/backward pass size (MB): 35.76
Params size (MB): 907.05
Estimated Total Size (MB): 943.68
----------------------------------------------------------------
```

Fig. 4: Variational Autoencoder architecture.

Unlike traditional autoencoders, VAEs combine the reconstruction probability, which measures how well the decoder can reconstruct the input data $x$ from the latent variable $z$

$$\text{Reconstruction Probability} = \mathbb{E}_{q_\phi(z|x)}[\log p_\theta(x|z)] \quad (1)$$

and the Kullback-Leibler (KL) divergence:

$$\text{KL Divergence} = \text{KL}(q_\phi(z|x) \parallel p(z)) \quad (2)$$

which measures the difference between the encoder's approximation $q_\phi(z|x)$ and the prior distribution $p(z)$, which, in our case, encourages the latent space to follow a standard normal distribution $N(0,1)$. Together, the VAE loss function is

$$L = \mathbb{E}_{q_\phi(z|x)}[\log p_\theta(x|z)] - \text{KL}(q_\phi(z|x) \parallel p(z)) \quad (3)$$

where $\phi$ represents the parameters of the encoder network, $p_\theta(x|z)$ is the decoder's likelihood of reconstructing the input $x$ given the latent variable $z$, and $\theta$ represents the parameters of the decoder network. By leveraging the loss function, we used the reconstruction probability to determine the presence of anomalies within each frame as described in [18], where VAE's outputs with high reconstruction error are labeled as anomalies and vise versa. The VAE's output is mapped onto [-1, +1] outputs. Using an experimentally determined threshold, if the reconstruction probability falls below this threshold, the output is set to +1, implying no anomaly; otherwise, it is set to -1, implying anomaly.

*D. OCSVM*

Another common approach for anomaly detection is the use of SVMs, as they can learn a boundary between recognized and non-recognized data [19]–[21]. However, OCSVMs struggle with high-dimensional data due to their high computational cost and extensive memory demands. By leveraging the VAE's latent space, which is a 1024element vector, we are able to train an OCSVM to learn a classification boundary to distinguish the known anomalies.

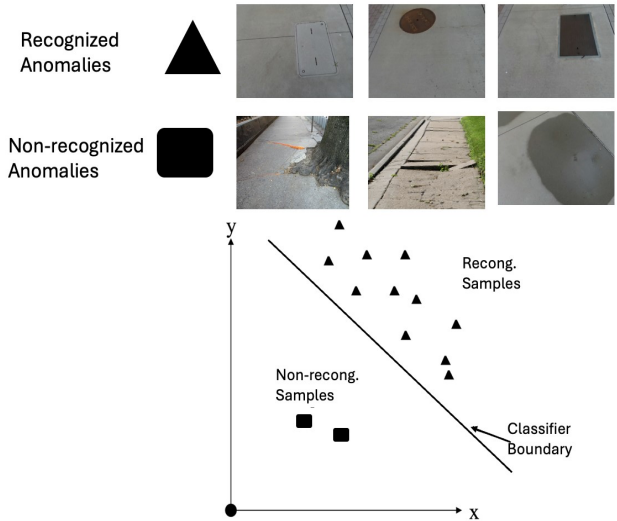

Fig. 5: Support Vector Machine Class Distinction.

To train the OCSVM, only one class of data was made available for training [22], as shown in Fig. 5. The data consisted of a set of unseen 640 x 480 images with the presence of non-hazardous anomalies, which were passed through the trained VAE and extracted as latent vectors. The data were then normalized between (-1,1) to ensure that all features contributed equally during the model's learning process. Although the latent vector is a small dimension, not every element in the vector is necessarily relevant. Thus, a Principal Component Analysis (PCA) was also used to further reduce the dimensionality of the latent vector, while maintaining the most important features, lowering the computational cost and memory requirements of the OCSVM.

Given the dimensionality and data distribution, a radial basis function (RBF) kernel with a scaling gamma ($\gamma$) was used. The OCSVM decision function returns +1 if the data is non-hazardous, otherwise it returns -1, as shown in equation 4

$$f(x) = \sum_{i=1}^{n} \alpha_i \cdot \exp\left(-\gamma \|x - x_i\|^2\right) - \mu \quad (4)$$

where $\alpha_i$ are the Lagrange multipliers, $\gamma$ is the scaling parameter for the RBF kernel, $\|x - x_i\|^2$ is the squared Euclidean distance, and $\mu$ is the bias term.

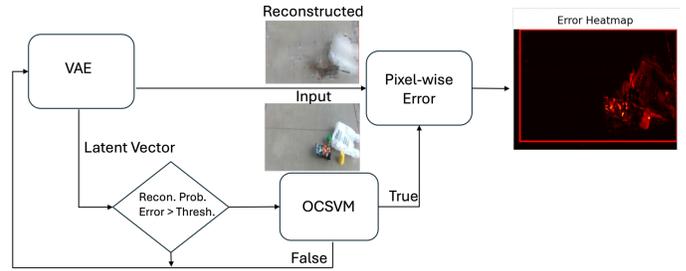

Fig. 6: Hybrid pipeline.

*E. Hybrid Output*

Sidewalk surfaces are not consistent. There are many infrastructure elements such as electrical boxes, manholes, and water covers that are not hazardous (i.e., they are safely traversable). However, when capturing data that was used to train the VAE, roughly 15% of the dataset consisted of these cases, impeding the VAE to learn meaningful features from these cases. As a result, the VAE will struggle to generate accurate reconstructions when these objects are present. These high reconstruction errors will cause the system to alert excessively, which is neither practical nor safe. Therefore, to overcome these false alerts, the results from the VAE are merged with the OCSVM.

When the VAE produces a -1 output, the corresponding PCA of the latent vector is passed to the OCSVM, which then determines if the input represents a hazard or a nonhazard. The decision function from the OCSVM outputs two discrete values [-1, +1] as well. If the OCSVM also outputs -1, indicating a hazard, a pixel-wise mean squared error between to generate a heat map of the hazardous object, as illustrated in Fig. 6. This indicates the location of the anomaly in the image frame. This is useful for diagnostic purposes and could be integrated into a navigation guidance system, where a bounding box is created to indicate the region where the hazardous obstacle is located.

Thus, once the VAE determines the presence of an anomaly, the latent vector is sent to an OCSVM to determine if the anomaly is hazardous or not. This method allows the system to ignore false hazard anomalies, such as the objects mentioned above.

## III. RESULTS

### A. Dataset

This dataset consists of more than 15,000 training frames and over 5,000 testing frames. Each frame is manually annotated as hazardous or non-hazardous. Within the entire dataset, there are three classes, as shown in Fig. 7. Additionally, there are seven examples of anomalies, i.e., objects on or in the sidewalk, but the system is not limited to these examples. To the best of our knowledge, this is the first dataset collected for the anomaly detection task in a sidewalk scenario.

- Case 1:
  - Not Anomalous

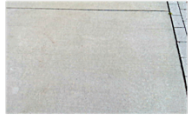

- Case 2:
  - Anomalous but not hazardous

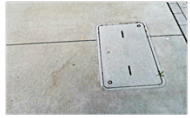

- Case 3:
  - Anomalous and hazardous

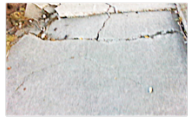

Fig. 7: Cases of objects within the dataset.

**Types of Hazardous Anomalies:**

- Puddle
- Gravel
- Uneven/broken sidewalk
- Pothole
- Nails
- Tree roots
- Litter or Debris

### B. Variational Autoencoder

The Receiver Operating Characteristic (ROC) curve illustrates the performance of the VAE alone, as shown in Figure 8. The ROC uses thresholds ranging from 10 to 500. Based on this ROC, the VAE performs well in detecting anomalies, with a strong ability to distinguish between true anomalies and false anomalies. The high Area Under the Curve (AUC) of 0.94 confirms the model's effectiveness.

### C. OCSVM

To tune the OCSVM, we used a subset of the dataset where only non-hazardous anomalies were present. We set the scaling gamma and $\mu$ of 0.5 with a radial basis function (RBF) kernel to obtain a classification accuracy of 90%.

| Predicted \ Actual | -1 | +1 |
|---|---|---|
| -1 | True Non-Hazard 2428 | False Hazardous 226 |
| +1 | False Non-Hazard 141 | True Hazardous 1031 |

TABLE I: Confusion Matrix for OCSVM only.

| Predicted \ Actual | -1 | +1 |
|---|---|---|
| -1 | True Non-Hazard 2465 | False Hazardous 189 |
| +1 | False Non-Hazard 141 | True Hazardous 1031 |

TABLE II: Confusion Matrix for hybrid (VAE + OCSVM).

### D. Hybrid Output

To validate the concept of hybrid VAE + OCSVM integration, we performed two experiments. The first experiment involved using only the VAE with a low threshold, while the second experiment incorporated the OCSVM with the VAE at the same threshold. Using the same validation data, where three cases are present (no anomalies, non-hazard anomalies, and hazard anomalies), we ran the algorithms to generate a set of confusion matrices, with 1 indicating hazards and -1 indicating non-hazards.

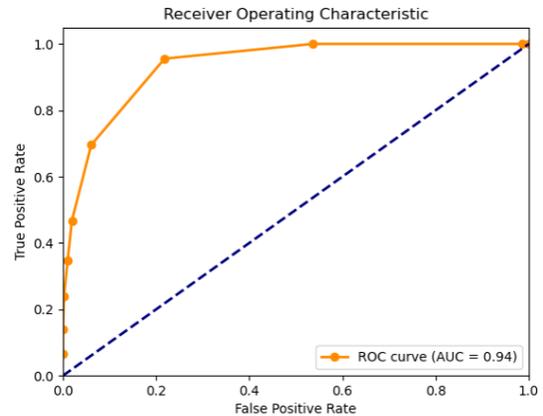

Fig. 8: Receiver Operating Characteristic Curve for proposed Variational Autoencoder

Table I shows the results for the VAE only, and Table II presents the results for the hybrid system. The number of false hazards decreases from 226 to 189, approximately 16%, demonstrating that the hybrid model achieves greater precision in distinguishing between hazardous and non-hazardous anomalies. This improvement results in a 91.4% accuracy for true hazard anomaly detection, highlighting the effectiveness of incorporating the OCSVM alongside the VAE.

IV. Discussion

This research introduces a system that employs visual data to enhance safety navigation by detecting hazardous objects within the trajectory of a sidewalk. From the ROC curve, we observe high accuracy with an AUC of 94% based on a range of thresholds. Opting for a high threshold leads to an increase in false non-hazard detections, representing the worst-case scenario. To minimize the rate of false non-hazards, selecting a lower threshold is optimal. However, this approach will increase the detection of false hazard situations, such as manholes, and water valves, among others. Consequently, the system's architecture, employing a variational autoencoder followed by a one-class SVM, effectively achieves strong performance and operational flexibility with an accuracy of 91.4%. In addition to the results provided, a supplementary video can be found to demonstrate the implementation of this work. The video demonstrate the three cases mentioned in Fig. 7 with indications of what the system is outputting. The system was designed for low computational cost during deployment. The autoencoder is able to perform well with low resolution images (640 x 480) and the OCSVM uses a small dimensional input as well, allowing the system to perform in real-time on a consumer laptop.

The VAE is used to distinguish normal sidewalk conditions from anomalies, while the OCSVM distinguishes between known anomalies and potential hazards. This method enables the VAE to be trained on extensive datasets prior to deployment, keeping the computational demands during real-time operation minimal. Moreover, given the lightweight computational requirements and low training cost of the OCSVM, it could be enhanced during deployment to recognize and differentiate repeated non-hazardous patterns. For instance, if a user is alerted to the same type of unrecognized anomaly but consistently navigates over it, the system could adapt by reclassifying that anomaly as non-hazardous.

While this study serves as a proof of concept for the system architecture, the data requirements for adequate training of the system for widespread deployment are not clear. We attempted to acquire data on various types of concrete sidewalks, and fortunately, due to the harsh winter weather in the Boston area, a wide variety of degraded sidewalk conditions are available. One challenge will be extending the system to non-concrete sidewalks, such as cobblestone streets in historic city centers. Here, the visual texture from the camera may make anomaly detection problematic. One potential solution is the incorporation of depth information from stereo cameras. This would also facilitate the detection of raised obstacles and concavities such as potholes. Depth information could be readily incorporated into the same architecture demonstrated here, albeit with increased computational demands. Additionally, the system's performance can be impacted on bright days when shadows obscure the area where the anomaly is present. Future plans include incorporating shadow removal methods to overcome regions darkened by shadows [23].

In conclusion, the key to the system's success is the use of RGB frames in combination with machine learning [24]. The systems' approach can increase independence when navigating environments where uncertainty is high. While this research primarily focuses on enhancing safety for the visually impaired, there are numerous potential applications beyond this scope. The system could provide crucial assistance in navigation for assistive lower-limb devices, controlling motorized wheelchairs, and legged devices, such as bipedal robots and prosthetics.

References

[1] A. Anwar, "A Smart Stick for Assisting Blind People," *IOSR Journal of Computer Engineering*, vol. 19, no. 3, pp. 86–90, 5 2017.

[2] P. Slade, A. Tambe, and M. J. Kochenderfer, "Multimodal sensing and intuitive steering assistance improve navigation and mobility for people with impaired vision," Tech. Rep., 2021. [Online]. Available: https://www.science.org

[3] H. Wang, J. Qin, A. Bastola, X. Chen, J. Suchanek, Z. Gong, and A. Razi, "VisionGPT: LLM-Assisted Real-Time Anomaly Detection for Safe Visual Navigation," 3 2024. [Online]. Available: http://arxiv.org/abs/2403.12415

[4] Z. Zou, K. Chen, Z. Shi, Y. Guo, and J. Ye, "Object Detection in 20 Years: A Survey," 5 2019. [Online]. Available: http://arxiv.org/abs/1905.05055

[5] Y. Falinie, A. Gaus, N. Bhowmik, B. K. S. Isaac-Medina, H. P. H. Shum, A. Atapour-Abarghouei, and T. P. Breckon, "Region-based Appearance and Flow Characteristics for Anomaly Detection in Infrared Surveillance Imagery," Tech. Rep.

[6] E. L. Andrade, S. Blunsden, and R. B. Fisher, "Modelling Crowd Scenes for Event Detection," Tech. Rep.

[7] H. Alsolai, F. N. Al-Wesabi, A. Motwakel, and S. Drar, "Assisting Visually Impaired People Using Deep Learning-based Anomaly Detection in Pedestrian Walkways for Intelligent Transportation Systems on Remote Sensing Images," *Journal of Disability Research*, vol. 2, no. 2, 8 2023.

[8] A. Tellaeche Iglesias, M. . Campos Anaya, G. Pajares Martinsanz, and I. Pastorlopez, "On combining convolutional autoencoders and support´ vector machines for fault detection in industrial textures," *Sensors*, vol. 21, no. 10, 5 2021.

[9] R. Gasparini, A. D'Eusanio, G. Borghi, S. Pini, G. Scaglione, S. Calderara, E. Fedeli, and R. Cucchiara, "Anomaly detection, localization and classification for railway inspection," in *Proceedings - International Conference on Pattern Recognition*. Institute of Electrical and Electronics Engineers Inc., 2020, pp. 3419–3426.

[10] L. Wellhausen, R. Ranftl, and M. Hutter, "Safe Robot Navigation via Multi-Modal Anomaly Detection," 1 2020. [Online]. Available: http://arxiv.org/abs/2001.07934http://dx.doi.org/10.1109/LRA.2020.2967706

[11] Derek Seward, Conrad Pace, and Rahee Agate, "Safe and effective navigation of autonomous robots in hazardous environments," *Springer Science + Business Media, LLC 2006*, 2006.

[12] E. Tricomi, M. Mossini, F. Missiroli, N. Lotti, X. Zhang, M. Xiloyannis, L. Roveda, and L. Masia, "Environment-Based Assistance Modulation for a Hip Exosuit via Computer Vision," *IEEE Robotics and Automation Letters*, vol. 8, no. 5, pp. 2550–2557, 5 2023.

[13] N. E. Krausz, T. Lenzi, and L. J. Hargrove, "Depth sensing for improved control of lower limb prostheses," *IEEE Transactions on Biomedical Engineering*, vol. 62, no. 11, pp. 2576–2587, 11 2015.

[14] A. H. Al-Dabbagh and R. Ronsse, "Depth Vision-Based Terrain Detection Algorithm During Human Locomotion," *IEEE Transactions on Medical Robotics and Bionics*, vol. 4, no. 4, pp. 1010–1021, 11 2022.

[15] K. Yang, K. Wang, R. Cheng, W. Hu, X. Huang, and J. Bai, "Detecting traversable area and water hazards for the visually impaired with a pRGB-D sensor," *Sensors (Switzerland)*, vol. 17, no. 8, 8 2017.

[16] P. Herghelegiu, A. Burlacu, and S. Caraiman, *Negative Obstacle Detection for Wearable Assistive Devices for Visually Impaired*, 2017.

[17] A. N. K, S. Ramachandran, N. George, and L. Shine, "Enhancing Outdoor Mobility and Environment Perception for Visually Impaired Individuals Through a Customized CNN-based System," Tech. Rep. 9. [Online]. Available: www.ijacsa.thesai.org

[18] J. An and S. Cho, "SNU Data Mining Center 2015-2 Special Lecture on IE Variational Autoencoder based Anomaly Detection using Reconstruction Probability," Tech. Rep., 2015.

[19] A. C. Braun, U. Weidner, and S. Hinz, "Classification in highdimensional feature spaces-assessment using SVM, IVM and RVM with focus on simulated EnMAP data," *IEEE Journal of Selected Topics in Applied Earth Observations and Remote Sensing*, vol. 5, no. 2, pp. 436–443, 2012.

[20] C.-W. Hsu, C.-C. Chang, and C.-J. Lin, "A Practical Guide to Support Vector Classification," Tech. Rep., 2003. [Online]. Available: http://www.csie.ntu.edu.tw/~cjlin

[21] D. Liu, H. Qian, G. Dai, and Z. Zhang, "An iterative SVM approach to feature selection and classification in high-dimensional datasets," *Pattern Recognition*, vol. 46, no. 9, pp. 2531–2537, 9 2013.

[22] B. Scholkopf, R. Williamson, A. Smola, J. Shawe-Taylor, J. Platt, and R. Holloway, "Support Vector Method for Novelty Detection," Tech. Rep.

[23] Z. Liu, H. Yin, X. Wu, Z. Wu, Y. Mi, and S. Wang, "From Shadow Generation to Shadow Removal," Tech. Rep. [Online]. Available: https://github.

[24] A. I. Khan and S. Al-Habsi, "Machine Learning in Computer Vision," in *Procedia Computer Science*, vol. 167. Elsevier B.V., 2020, pp. 1444–1451.